\documentclass[10pt,twocolumn,letterpaper]{article}

\usepackage[accsupp]{axessibility}  
\usepackage{iccv}
\usepackage{times}
\usepackage{epsfig}
\usepackage{graphicx}
\usepackage{amsmath}
\usepackage{amssymb}
\usepackage{multirow}
\usepackage{makecell}
\usepackage{subcaption}
\usepackage{enumitem}
\usepackage[pagebackref=true,breaklinks=true,letterpaper=true,colorlinks,bookmarks=false]{hyperref}

\long\def\th{$^{th}\enspace$}

\DeclareMathOperator{\myAE}{AE}
\DeclareMathOperator{\MAE}{MAE}
\DeclareMathOperator{\AMAE}{AMAE}
\iccvfinalcopy 

\def\iccvPaperID{3963} 

\ificcvfinal\fi

\begin{document}

\title{Ground-truth or DAER: Selective Re-query of Secondary Information}

\author{Stephan J. Lemmer and Jason J. Corso\\
University of Michigan, Ann Arbor\\
{\tt\small \{lemmersj, jjcorso\}@umich.edu}
}

\maketitle
\ificcvfinal\thispagestyle{empty}\fi

\begin{abstract}
Many vision tasks use secondary information at inference time---a seed---to assist a computer vision model in solving a problem.  For example, an initial bounding box is needed to initialize visual object tracking. To date, all such work makes the assumption that the seed is a good one.  However, in practice, from crowdsourcing to noisy automated seeds, this is often not the case.  
We hence propose the problem of seed rejection---determining whether to reject a seed based on the expected performance degradation when it is provided in place of a gold-standard seed. We provide a formal definition to this problem, and focus on two meaningful subgoals: understanding causes of error and understanding the model's response to noisy seeds conditioned on the primary input. With these goals in mind, we propose a novel training method and evaluation metrics for the seed rejection problem. We then use seeded versions of the viewpoint estimation and fine-grained classification tasks to evaluate these contributions. In these experiments, we show our method can reduce the number of seeds that need to be reviewed for a target performance by over 23\% compared to strong baselines.
\end{abstract}
%
\begin{figure}
    \centering
    \includegraphics[width=0.72\linewidth]{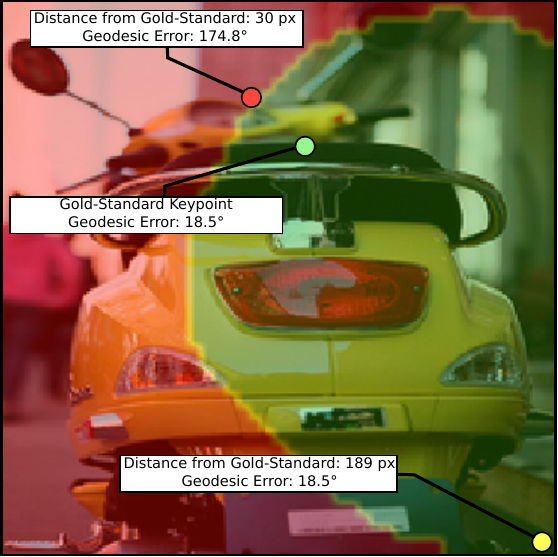}
    \caption{An example from keypoint-conditioned viewpoint estimation~\cite{szeto_click_2017}, with a heatmap of error caused by all potential clicks overlaid. Approaches focused on input-space accuracy~\cite{song_c-reference_2020, song_popup:_2019, dai_pomdp-based_2013, russakovsky_best_2015, deng_imagenet_2009, real_youtube-boundingboxes_2017, lin_microsoft_2014} would select the red keypoint over the yellow keypoint as it closer to the gold-standard (green) keypoint, even though this results in higher error.}
    \label{fig:unintuitive_error}
\end{figure}
\section{Introduction}
Many tasks in computer vision require not only a \textit{primary input}, such as an image or a video, but also additional information based on the primary input---a \textit{seed}---to be provided to the task model. This seed may be used to define the problem, such as in visual object tracking~\cite{kristan_novel_2016}, video object segmentation~\cite{perazzi_benchmark_2016}, and visual question answering~\cite{antol_vqa:_2015}, or to provide additional information for common tasks such as fine-grained scene classification~\cite{koperski_plugin_2020}, visual concept prediction~\cite{wang_feedback-prop_2018}, or viewpoint estimation~\cite{szeto_click_2017}. Critically, these tasks are evaluated using verified gold-standard seeds, ignoring the noisy processes by which seeds are generated. 

The performance of computer vision models with poor primary inputs has been explored in the context of naturally difficult~\cite{vasiljevic_examining_2017,zhou_classification_2017,dodge_understanding_2016,geifman_selectivenet_2019} and intentionally adversarial~\cite{wiyatno_physical_2019,wu_making_2019,eykholt_robust_2018,szegedy_intriguing_2014} primary inputs, leading to a variety of methods designed to make models more robust~\cite{vasiljevic_examining_2017, zhou_classification_2017} or detect and reject difficult inputs~\cite{geifman_selectivenet_2019}. However, no work to our knowledge has been performed on the identification and rejection of \emph{bad seeds}: seeds that cause a significant increase in error on the task when used in place of the gold-standard seed. As reliability issues in crowdsourcing are well studied~\cite{ipeirotis_quality_2010,raykar_eliminating_2012,song_two_2018,rzeszotarski_instrumenting_2011} and automated systems that could be used to create seeds are subject to unpredictable failure modes~\cite{rosenfeld_elephant_2018, wu_making_2019}, not having any mechanism for detecting bad seeds is a critical oversight.

To emphasize the need for such a mechanism, we examine Figure~\ref{fig:unintuitive_error}, where a human annotator is asked to click a semantically meaningful location on the image (\eg rear seat) to resolve the viewpoint estimation model's perceptual ambiguities. This example illustrates the complex, and sometimes counterintuitive, interaction between the primary input, seed, and task model: while many seeds that are incorrect in the input space (\eg the yellow seed) don't degrade the performance, many that are nearly correct in the input space (\eg the red seed) perform significantly worse than the gold-standard. It follows that current methods, which are designed to optimize accuracy for dataset curation~\cite{song_popup:_2019,dai_pomdp-based_2013, russakovsky_best_2015, deng_imagenet_2009, real_youtube-boundingboxes_2017, lin_microsoft_2014}, are inadequate for this task not only because they require excessive additional seeds to achieve consensus, but also because they optimize the wrong objective: they maximize accuracy in the input space at the potential cost of output accuracy.
\begin{figure}
    \centering
    \includegraphics[width=\linewidth]{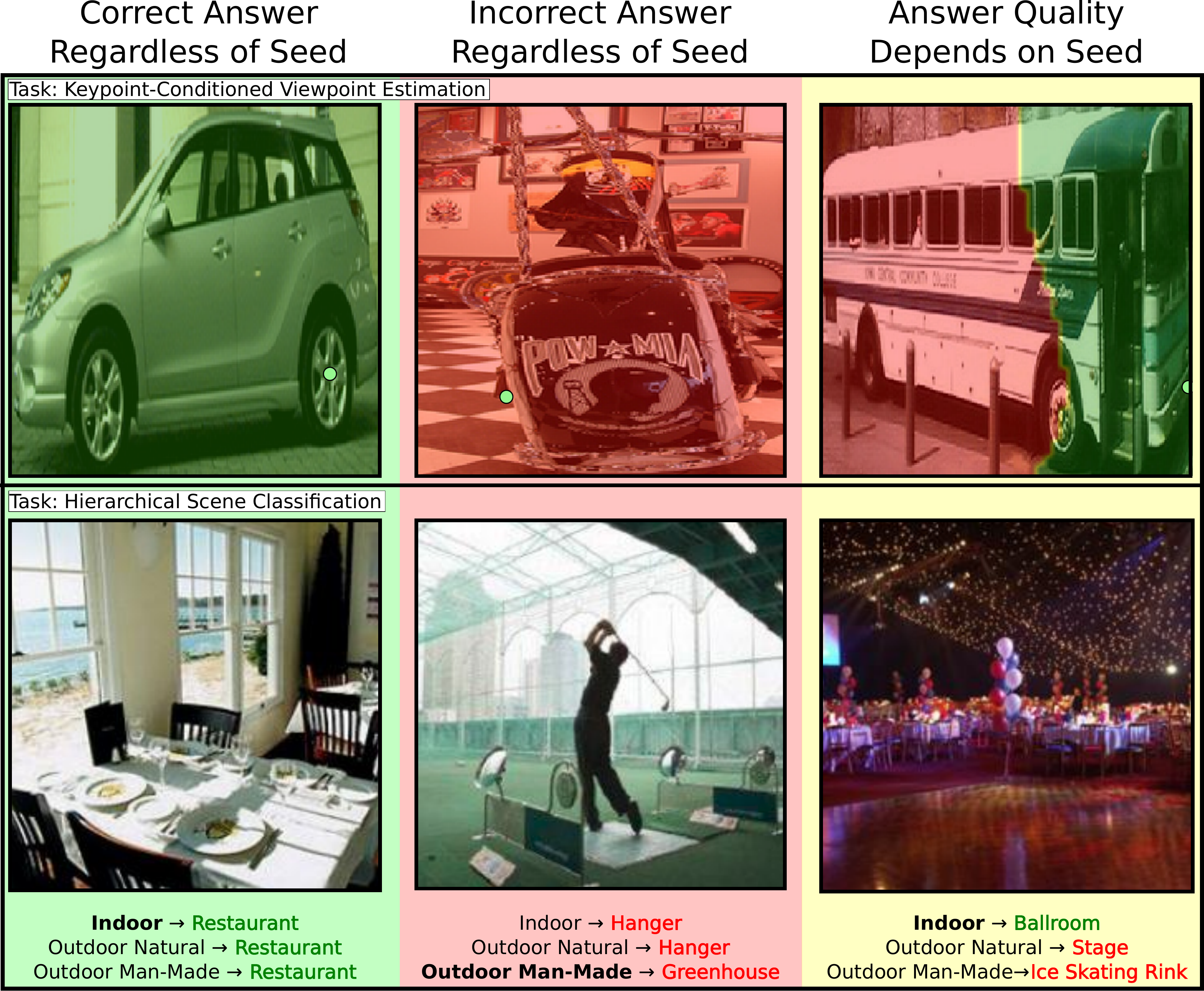}
    \caption{On both the KCVE (top row) and HSC (bottom row) tasks, the task model may or may not condition its answer solely on the primary input. For KCVE, the gold-standard seed is shown as a green circle, while the overlaid heatmap shows error from low (green) to high (red). For HSC, the gold-standard seed is in bold, correct answers are shown in green, and incorrect answers are shown in red.}
    \label{fig:seed_effects}
\end{figure}

In this work, we resolve this critical oversight by directly studying the problem of \textit{seed rejection}. Seed rejection seeks a principled mechanism for discarding candidate seeds that result in a less accurate output than the corresponding (unknown at inference time) gold-standard seed. A rejected seed could then be requeried to improve overall performance. Conceptually, we separate the problem of seed rejection into two distinct sub-goals: 

\noindent\textbf{Understanding the Cause of Error}: The first goal is understanding the degree to which the seed affects the output of the task model. If the seed has no influence on the task model's output (Figure~\ref{fig:seed_effects}, left and center), requesting another seed would be of little benefit. While the task of selective prediction~\cite{geifman_selectivenet_2019, chow_optimum_1970} has been proposed for handling bad primary inputs, no work to our knowledge has been performed on the task of rejecting bad seeds independent of the primary input's quality.

\noindent\textbf{Understanding the Task Model Response}: Next, we must gain an understanding of the task model's response, and how a human's intuition of a seed's quality differs from its effect on the accuracy of the task model's output. We again highlight the example shown in Figure~\ref{fig:unintuitive_error}, where a small Euclidean error in the input space (red keypoint) can cause a large increase in output error, while a much larger Euclidean error (yellow keypoint) may have little effect.

To address these challenges, we propose Dual-loss Additional Error Regression (DAER), a novel training method developed for the seed rejection problem. DAER considers the two challenges discussed above separately during training, and combines them during inference to predict the effect of a candidate seed on the downstream task. We evaluate the performance of DAER on two tasks: keypoint-conditioned viewpoint estimation~\cite{szeto_click_2017}---a human-in-the-loop extension of the canonical viewpoint estimation task~\cite{tulsiani_viewpoints_2015, su_render_2015,ferrari_starmap_2018,mustikovela_self-supervised_2020,liao_spherical_2019}---and hierarchical scene classification~\cite{koperski_plugin_2020}---a method that improves performance on fine-grained classification~\cite{wah_caltech-ucsd_2010, zhou_places_2018,li_learning_2018,xiao_sun_2010} by integrating a coarse scene classification. 

To evaluate DAER, we introduce a task-agnostic benchmark evaluation method for seed rejection, centered around new metrics designed specifically to assess the performance of a seed rejection method: Additional Error (AE), Mean Additional Error (MAE), and Area under the Mean Additional Error curve (AMAE). Unlike existing metrics, such as selective risk~\cite{geifman_bias-reduced_2019}, these metrics focus on the potential benefit of a new seed, instead of an oracle label of the target value that may be prohibitively difficult to obtain at scale.

The contributions of this paper are as follows:
\begin{enumerate}[itemsep=0pt,itemindent=0pt,topsep=0pt,leftmargin=3ex,parsep=0pt]
    \item A formalization and benchmark metrics for the seed rejection problem, in which a model is tasked with determining if a candidate seed will produce significantly higher error than the corresponding (unknown at inference time) gold-standard seed.
    \item Dual-loss Additional Error Regression (DAER), a broadly applicable training and inference method for the task of seed rejection. 
    \item An evaluation of DAER on the tasks of keypoint-conditioned viewpoint estimation~\cite{szeto_click_2017} (KCVE) and hierarchical scene classification~\cite{koperski_plugin_2020} (HSC), which shows that DAER can reduce the the number of seeds that need to be reviewed for a given target performance by over 23\% compared to the best-performing baseline.
\end{enumerate}
\section{Related Work}
\subsection{Seeded Inference}
Seeded inference describes a number of problems in which a task model accepts a primary input and additional information based on that primary input---a seed---and estimates a target value. Though the list of problems that can be classified as seeded inference is long~\cite{song_c-reference_2020, branson_visual_2010, song_popup:_2019, griffin_video_2020, reddy_carfusion:_2018, wang_feedback-prop_2018, szeto_click_2017, koperski_plugin_2020, hu_segmentation_2016, perazzi_benchmark_2016, kristan_novel_2016, antol_vqa:_2015, gurari_vizwiz_2018}, this is the first work to explicitly consider them as a class of problems. 

While some work acknowledges that performance can be improved by choosing which seed to request~\cite{banani_adviser_2018, griffin_bubblenets_2019}, current work generally does not consider the seed itself to be subject to error. In cases where the seed is categorical, such as hierarchical scene classification~\cite{koperski_plugin_2020, wang_feedback-prop_2018}, seeds other than the gold-standard are not considered. In contrast, many works in which the input space is effectively continuous, such as keypoint clicks in keypoint-conditioned viewpoint estimation~\cite{szeto_click_2017} and bounding boxes in visual object tracking~\cite{kristan_novel_2016, wu_online_2013}, acknowledge that seeds can be noisy and either seek to improve robustness~\cite{shah_cycle-consistency_2019}, or simply evaluate the robustness of existing models to a range of expected noise defined a-priori~\cite{wu_online_2013, szeto_click_2017}. Critically, in addition to ignoring the effect of seeds that are not within this predefined range, these methods do not consider which specific seeds result in an increase in error.

%
\subsection{Selective Prediction}
\label{sec:selective_prediction}
A problem closely related to seed rejection is the problem of selective prediction \cite{chow_optimum_1970, geifman_selectivenet_2019}. In selective prediction, the goal is to split primary inputs into a set that is classified by a task model and a set that is classified by expert human annotators such that annotation cost is minimized subject to an error constraint, or error is minimized subject to a cost constraint. Selective prediction has been applied to many regression and classification strategies over time, from nearest neighbors in the 1970's~\cite{hellman_nearest_1970}, to support vector machines in the early 2000s~\cite{goos_support_2002}, to deep artificial neural networks today~\cite{yildirim_leveraging_2019, geifman_selectivenet_2019}. Gurari \etal~\cite{gurari_pull_2016} extend the problem of selective prediction by considering the case where multiple models (including human annotators) are available, and predicting a best performer based on a regressed intersection-over-union.

While both selective prediction and seed rejection predict performance of a task model on a given input, selective prediction only considers rejection of a single input, which would be seen in tasks such as as image classification~\cite{geifman_selective_2017, raghu_direct_2019, geifman_bias-reduced_2019} or tabular regression~\cite{goos_optimal_2000, geifman_selectivenet_2019}. In these tasks, the only option if the primary input is rejected is to receive a target label from a human expert. This results in an unnecessary increase in annotation cost due to the target label being inherently more difficult to obtain than a seed. For example, it is substantially easier to perform a keypoint click than a full viewpoint annotation~\cite{szeto_click_2017}, or to initialize an object tracker with a first-frame bounding box than draw a bounding box on every video frame~\cite{kristan_novel_2016}. 
\section{Seed Rejection}
\label{sec:problem_definition}
Here, we first define seed rejection and its associated metrics in a problem-independent manner, where task and rejection models may be parameterized by learned or hard-coded methods. Next, we present a generic formulation of our proposed solution, which we call Dual-loss Additional Error Regression (DAER). In Section \ref{sec:experiments}, we instantiate this methodology in two concrete problems.
\begin{figure}[t]
    \centering
    \includegraphics[width=\linewidth]{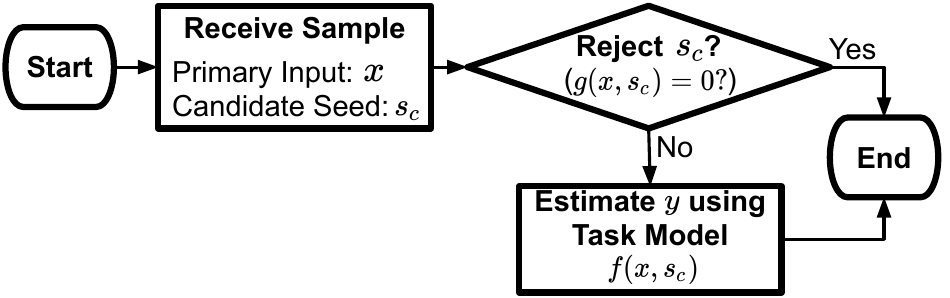}
    \caption{A flowchart of seed rejection on a single sample $(x, s_c) \in \mathcal{D}$. The rejection model, $g(x, s_c)\in \{0, 1\}$ seeks to reject samples for which using the candidate seed results in worse performance than using the (unknown at inference) gold-standard seed.}
    \label{fig:rejection_example}
\end{figure}
\begin{figure*}[t]
    \centering
    \includegraphics[width=0.8\linewidth]{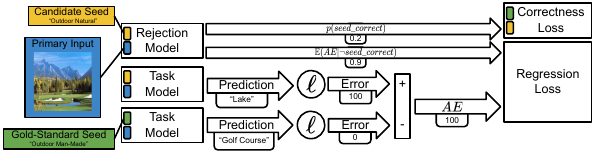}
    \caption{DAER separates the regression of additional error into two components: predicting whether the candidate seed is correct through a correctness loss, and predicting the additional error through a regression loss, which is only backpropagated if the candidate seed is incorrect. For illustration, we include an example from the hierarchical scene classification task.}
    \label{fig:model_aware_training}
\end{figure*}
\subsection{Problem Statement}
Seed rejection, shown in Figure~\ref{fig:rejection_example}, is based around a task model, $f(x, s)$, which accepts a primary input, $x$, and extra information based on that primary input---a seed, $s$---that is either a candidate, $s_c$, or gold-standard, $s_{gs}$, ($s \in \{s_c, s_{gs}\}$). Given these inputs, the task model provides an estimate of a target value, $y$, with the goal of minimizing a task-specific performance measure, $\ell$. 

We then consider how using a candidate seed in place of the gold-standard seed affects the inference-time output of a certain task model for a given primary input. While the gold-standard seed is unknown at inference time, during training and evaluation it acts as a verified ``true'' seed to compare the performance of a candidate seed to. We measure this change in performance using a new metric we call \textit{additional error} ($\myAE$), which is given as:
\begin{align}
    \myAE(&x,s_c,s_{gs},y|f,\ell) =\nonumber\\ &\max(\ell(f(x,s_c),y)-\ell(f(x,s_{gs}),y),0)\enspace.
\label{eq:additional_error}
\end{align}
Critically, we note the $\max$ operator, which enforces the constraint that a candidate seed cannot outperform the corresponding gold-standard seed. This is important in cases such as those shown in Figure~\ref{fig:reject_examples}-(B) and (C), where there exist seeds that perform better than the gold-standard, but we can not expect a method tasked with returning the gold-standard seed to provide it.

We seek a rejection model, $g(x, s_c) \in \{0, 1\}$, such that candidate seeds with low additional error are accepted ($g(x, s_c) = 1$), and candidate seeds with high additional error are rejected ($g(x, s_c) = 0$). While an ideal rejection model would be able to divide candidate seeds into ``correct'' and ``incorrect'' bins, in practice the goal is to optimize a tradeoff between the proportion of candidate seeds that are accepted (referred to as \textit{coverage}) and an aggregate measure of the task model's performance over the accepted set (Section~\ref{sec:aggregate}). This is for two reasons: first, such techniques are often subject to budgetary constraints, meaning the rejection model may need to accept candidate seeds that cause more error than the gold-standard seed but less than other candidate seeds (particularly in the case of continuous performance measures~\cite{szeto_click_2017, perazzi_benchmark_2016,kristan_novel_2016}). Next, the rejection model may be uncertain of the quality of the seed, and needs to balance its confidence with the cost of rejecting a seed. This suggests that seed rejection models should effectively rank primary input-seed pairs in order of desirability.

For example, in the case where keypoint-conditioned viewpoint estimation is used at scale to generate datasets via scene reconstruction~\cite{song_popup:_2019}, many workers would be asked to click specific keypoints on selected frames. A certain number of these keypoint clicks, targeting a budget or error tolerance through a method such as selective guaranteed risk~\cite{geifman_selective_2017}, are escalated to a more experienced worker who is assumed able to provide the correct seed.

\subsection{Aggregate Metrics}
\label{sec:aggregate}
With the goal of a rejection model defined, we note that aggregate metrics are required for parameter tuning and comparing the performance of rejection models on a test set, $\mathcal{D}$. 
We hence propose the Mean Additional Error (MAE), which corresponds to the mean of all additional errors across an accepted set of samples:
\begin{align}
    \MAE&(f,g|\mathcal{D},\ell) =
    \label{eq:mae}
    \\ &\frac{\frac{1}{|\mathcal{D}|}\sum_{(x,s_c,s_{gs},y)\in \mathcal{D}} {g(x,s_c) \myAE(x,s_c,s_{gs},y|f,\ell)}}{\frac{1}{|\mathcal{D}|}\sum_{(x, s_c)\in \mathcal{D}} g(x,s_c)}\enspace. \nonumber
\end{align}
%

Since a target coverage or MAE is chosen based on an application constraint (\eg, budget) we further seek a metric that can compare rejection models across all coverages. For this, we introduce the Area under the Mean Additional Error curve (AMAE) metric. This metric is found in two steps: first, we calculate the mean additional error at all coverages to produce a curve like the one shown in Figure~\ref{fig:AEHierarchical}. Next, we calculate the area under this curve. For a test set where the samples are ordered by the coverage where they are first accepted, this can be calculated empirically using the equation:
\begin{align}
    \AMAE = \frac{1}{|\mathcal{D}|}\sum_{i=1}^{|\mathcal{D}|} \frac{\sum_{j=1}^i \myAE(x^j,s_c^j,s_{gs}^j,y^j|f,\ell)}{i}\enspace.
    \label{eq:amae}
\end{align}
\noindent The AMAE can then be used to directly compare rejection models across all target coverages. For all proposed metrics (AE, MAE, AMAE), a lower value corresponds to a better performance.
\subsection{DAER}
\label{sec:method}
We approach the task of seed rejection by using a regressed estimate of the additional error (Equation~\ref{eq:additional_error}) as a scoring function to which a threshold can be applied. This regression is learned through a novel method we call \textit{Dual-loss Additional Error Regression} (DAER). Core to DAER is the separation of the additional error regression into two components corresponding to the challenges described in the introduction. The correctness loss, which addresses the subgoal \textit{understanding the cause of error}, is a classifier which estimates the likelihood that seed is correct. The regression loss, which addresses the subgoal \textit{understanding task model response}, estimates the additional error given that the seed is incorrect. That is, the regression loss is only used for training when the given seed is incorrect. This overall procedure is shown in Figure~\ref{fig:model_aware_training}.

Mathematically, the correctness and regression outputs can be used to calculate the expected additional error:
\begin{align}
    \mathbb{E}(\myAE(x^i, s_c^i, s_{gs}^i, y^i|f,\ell)) &=\\ p(\text{seed\_correct})&\mathbb{E}(\myAE | \text{seed\_correct}) \;+ \nonumber\\p(\neg \text{seed\_correct})&\mathbb{E}(\myAE | \neg \text{seed\_correct})\enspace. \nonumber
\end{align}
\noindent Since the additional error for a correct seed is always zero, this simplifies to:
\begin{align}
    \mathbb{E}(\myAE(x^i, s_c^i, s_{gs}^i, y^i|f,\ell)) &=\\ p(\neg \text{seed\_correct})&\mathbb{E}(\myAE | \neg \text{seed\_correct})\enspace. \nonumber
\end{align}
We use this formula to predict the additional error at inference time, but not during training. Instead, we train $p(\neg \text{seed\_correct})$ and $\mathbb{E}(\myAE | \neg \text{seed\_correct})$ with separate losses, a method that is the key component of DAER. While DAER's training method is mathematically equivalent to regressing the additional error directly, we show in Section~\ref{sec:subgoal_importance} that separating the two components significantly improves performance.
\section{Experiments}
\label{sec:experiments}
Our seed rejection method is applicable to a wide variety of problems, as it is fully specified by a four-tuple containing
a (fixed) task model, a rejection model architecture, a performance measure, and a definition of a correct seed. In this section, we demonstrate this flexibility by showing state-of-the-art performance on two disparate tasks: keypoint-conditioned viewpoint estimation and hierarchical scene classification. Extra details on training and evaluation for both tasks are available in our supplementary material and code repository\footnote{\href{https://github.com/lemmersj/ground-truth-or-daer}{https://github.com/lemmersj/ground-truth-or-daer}}. 
\subsection{Keypoint-Conditioned Viewpoint Estimation}
\label{sec:kcve}
Keypoint-conditioned viewpoint estimation~\cite{szeto_click_2017} is a human-in-the-loop extension of the canonical computer-vision task of viewpoint estimation~\cite{tulsiani_viewpoints_2015, su_render_2015,ferrari_starmap_2018,mustikovela_self-supervised_2020,liao_spherical_2019}. In this task, a human annotator is given an image of a vehicle, and asked to click a keypoint such as ``front right tire.'' This human-produced information is then combined with features from a convolutional neural network to estimate the camera viewpoint more accurately than would be possible without the keypoint~\cite{su_render_2015, tulsiani_viewpoints_2015}. 

In this work, we use the Click-Here CNN architecture~\cite{szeto_click_2017} as our task model and, with modified output layers, our rejection model. For evaluation, our performance measure is the geodesic on the unit sphere, following convention~\cite{szeto_click_2017, su_render_2015, tulsiani_viewpoints_2015}. However, it is impractical to use this measure during training due to the computational difficulty of calculating the matrix logarithm. Instead, our rejection model predicts rotational displacement in terms of Larochelle~\etal's distance~\cite{larochelle_distance_2007},
\begin{align}
    d = ||I-A_2A_1^T||_F \enspace ,
\end{align}
\noindent where $A_1$ and $A_2$ are the rotation matrices produced by the ground-truth and regressed Euler angles.

While it is intuitive to define a correct seed as one that exactly matches the gold-standard seed, the Click-Here CNN architecture uses a 46x46 one-hot grid as a seed, which makes it unlikely that a randomly selected point will match the gold-standard keypoint. Therefore, defining a correct seed in this way would result in a rejection model whose objective effectively reduces to regressing the additional error directly.
Instead, we define a correct seed as a seed for which the additional error is zero:
\begin{align}
p(\text{seed\_correct}) = 
\begin{cases} 
      0 & \myAE = 0 \\
      1 & \myAE \neq 0 
\end{cases}\enspace.
\end{align}
\noindent In addition to more effectively balancing correct and incorrect seeds, defining a correct seed in this way encourages the rejection model to take a shortcut by learning the interaction between the task model and primary input prior to considering the seed. For example, the left and center cases in Figure~\ref{fig:seed_effects} can be accepted without considering the location of the seed.

\noindent\textbf{Training}
%
%
During training, candidate seeds are generated by randomly sampling a pixel within the input image crop. For the correctness loss, we use binary cross-entropy, while we follow the common convention of using binned cross-entropy for the regression loss~\cite{tulsiani_viewpoints_2015, su_render_2015}.

\noindent\textbf{Evaluation} %
We maintain the human-in-the-loop motivation of the original work by evaluating with crowdsourced keypoints for our seeds. We collected a total of 6,042 keypoints on the PASCAL3D+ validation set~\cite{xiang_beyond_2014} from US-based annotators via Amazon Mechanical Turk. 
In order to produce a representative seed distribution for validation, we divide the PASCAL3D+ validation set and corresponding crowdsourced seeds into five folds such that no vehicle crop appears in more than one fold, and report the mean across folds.

\noindent\textbf{Baselines}
Our baselines for seed rejection on the keypoint-conditioned viewpoint estimation task are:
\begin{itemize}[topsep=0pt,itemsep=-1ex,partopsep=1ex,parsep=1ex,itemindent=-2ex]
\item Softmax Response (S.R.): The largest value of the softmax output. This was shown by Geifman \& El-Yaniv~\cite{geifman_selective_2017} to perform best on selective prediction, the task most similar to seed rejection. 
\item Known Distance: Oracle knowledge of the candidate seed's Euclidian distance from the gold-standard seed. This has a relation to crowdsourcing approaches, which seek to minimize error in the input space.
\item Task Network Entropy: The distributional entropy of the output of the task model. 
\item Task Network Percentile: 10,000 samples are taken from the task model's output distribution, and the 80\th percentile difference between all samples and the mean is used as our rejection criteria. Results for other percentiles are given in supplementary material.
\end{itemize}

\begin{figure}[t]
    \centering
    \includegraphics[width=0.75\linewidth]{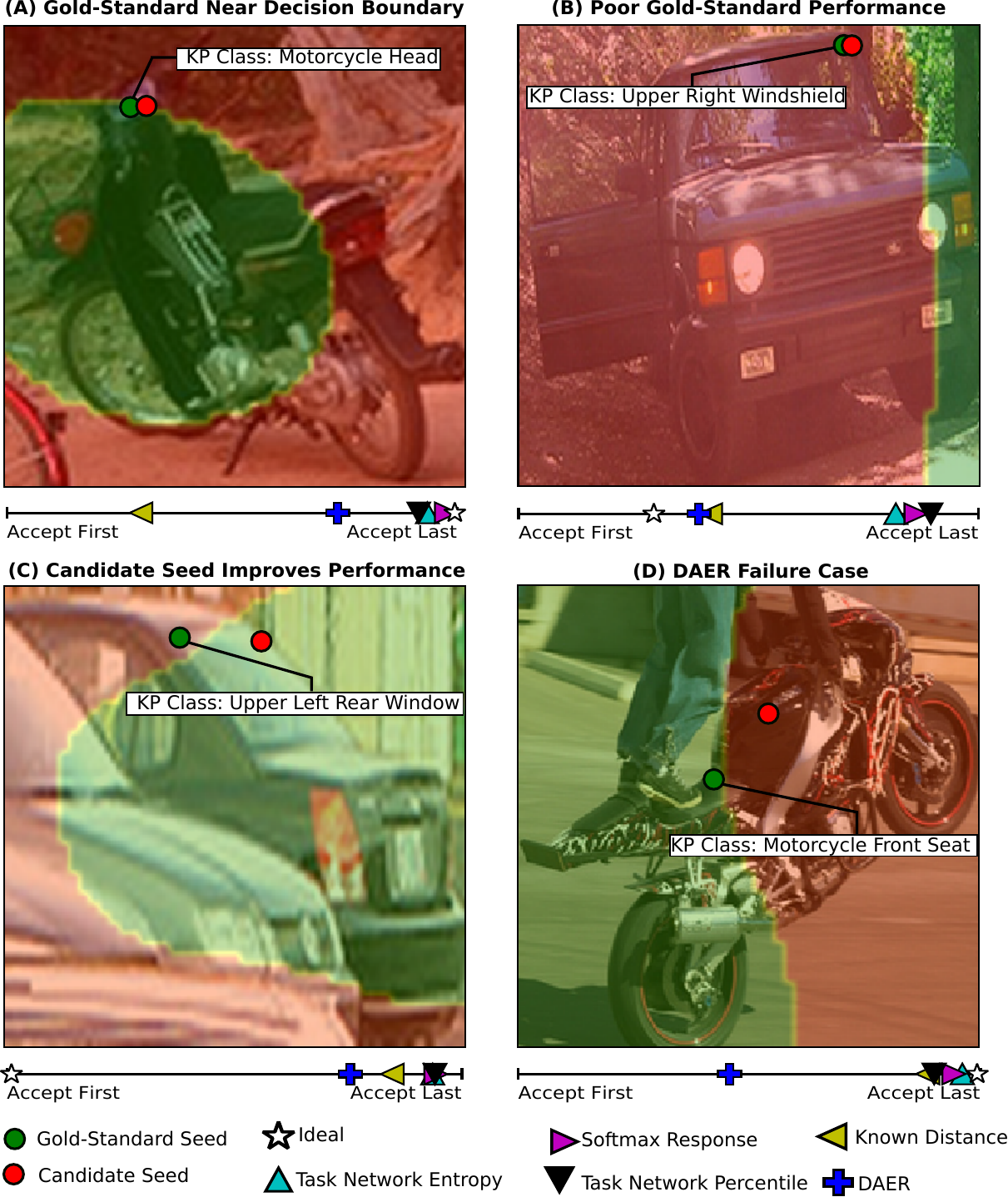}
    \caption{Select example cases from KCVE. Ideal accept location---the coverage at where sorting by additional error would accept a seed---is given by the white star. Overlaid heatmaps are from green (low error) to red (high error)}
    \label{fig:reject_examples}
\end{figure}

\noindent\textbf{Results}
We show in Table~\ref{tab:kcve_performance} that DAER outperforms baselines on the keypoint-conditioned viewpoint estimation task. We highlight specific examples in Figures~\ref{fig:reject_examples} and \ref{fig:regressed_AE}. In \ref{fig:reject_examples}-(A), we see an extreme case where the gold-standard is near the decision boundary and there is a high additional error even though the candidate seed is near the gold-standard seed. This causes the known distance baseline to fail by accepting the candidate seed early, while DAER and baselines based on the task model's output recognize a high probabilty of error and accept this candidate seed late. In \ref{fig:reject_examples}-(B), we highlight a case where DAER successfully recognizes that while the geodesic error for the candidate seed is high, the ground-truth seed will not provide an improved estimate of the camera viewpoint. \ref{fig:reject_examples}-(C) represents a similar case in which the gold-standard seed causes error in the output, but in this case the candidate seed produces a better output, despite a mismatch between the keypoint label and location. In \ref{fig:reject_examples}-(D), we see a failure case, where DAER is unable to accurately estimate the task model's decison boundary, resulting in early acceptance of a poor seed. 

\begin{table}[t]
\centering
\begin{tabular}{|l|l|}
\hline
Method           & AMAE            \\ \hline
Random & 1.54          \\ \hline
Softmax Response & 0.9306          \\ \hline
Known Distance   & 0.3964          \\ \hline
Task Network Entropy          & 0.3534          \\ \hline
Task Network Percentile       & 0.3092          \\ \hline
DAER             & \textbf{0.2864} \\ \hline
\end{tabular}%
\caption{Mean AMAE for baselines and DAER across all folds on the KCVE task (lower is better).}
\label{tab:kcve_performance}
\end{table}%

\subsection{Hierarchical Scene Classification}
Hierarchical scene classification~\cite{hu_learning_2016,koperski_plugin_2020,wang_feedback-prop_2018} is an extension of fine-grained classification~\cite{wah_caltech-ucsd_2010,zhou_places_2018,li_learning_2018,xiao_sun_2010} in which information about the coarse scene categorization---such as ``indoor''---is given to a classifier alongside the image to help determine the fine-grained scene classification---such as ``ballroom''---of an image. 
In this work, we train and evaluate on the SUN397 dataset~\cite{xiao_sun_2010}, a dataset of over 130,000 images across 397 classes, and use the Plugin Network architecture developed by Koperski \etal~\cite{koperski_plugin_2020} as our task model.
For this problem, we define the correct seed as the seed that matches the gold-standard coarse classification. The performance measure is given as:
\begin{align}
\ell(f(x,s), y) =
\begin{cases} 
      0 & f(x, s_c) = y \\
      100 & f(x, s_c) \neq y
\end{cases}\enspace.
\end{align}
\noindent With this performance measure, the MAE corresponds to the percent difference in accuracy caused by using candidate seeds in place of the gold-standard seeds at a given coverage.

\noindent\textbf{Training}
The rejection model, based on a pretrained ResNet-18 architecture, is trained using a randomly selected coarse category as the seed, and validation is performed using all potential seeds for a primary input. The instance of the rejection model with the lowest validation AMAE is used for testing. Full training details are available in the supplementary material.

\noindent\textbf{Evaluation}
For the hierarchical scene classification task, the seed is produced via a classification model trained to predict one of the 7 coarse category combinations (details in supplementary). We train five rejection models and five seeding models. This allows us to calculate the standard error across 5 runs for the baseline methods, and across 25 runs for the learned rejection models.

\noindent\textbf{Baselines}
As our baselines, we use the task network entropy and softmax response scores described in Section~\ref{sec:kcve}. Since the seed is provided by a DNN classifier, we apply these baselines to both the output of the task model, which we prefix with the term ``fine'', and the output of the seeding model, which we prefix with the term ``coarse''.

\begin{figure}[b]
    \centering
    \includegraphics[width=.8\linewidth]{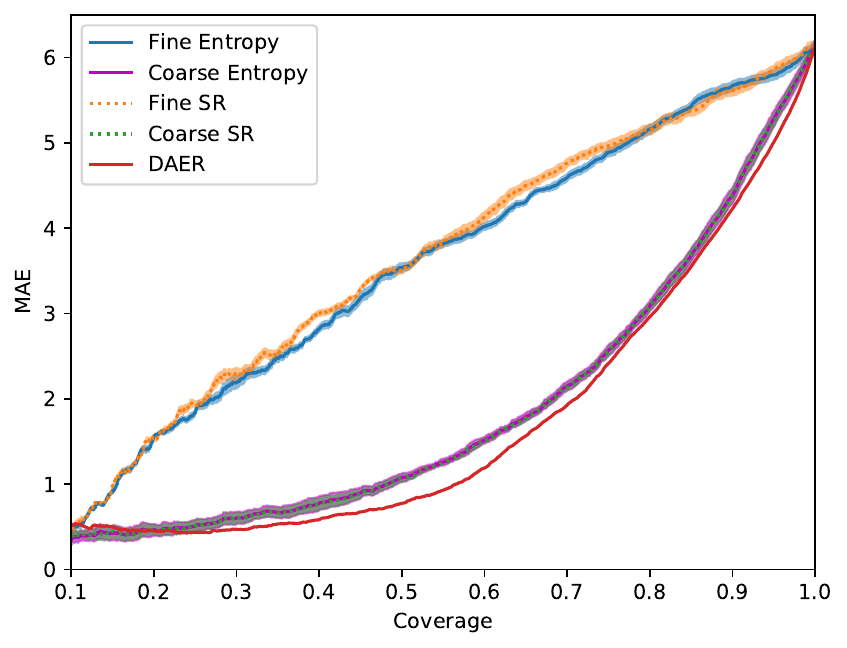}
    \caption{The mean additional error compared to the proportion of seeds accepted  (coverage) for the hierarchical scene classification task (lower is better). The dark lines represent the mean of all runs. The shaded area represents one standard error.}
    \label{fig:AEHierarchical}
\end{figure}

\noindent\textbf{Results}
We see in Table~\ref{tab:HSC_performance} that DAER significantly outperforms baselines on the seed rejection task for hierarchical scene classification under the aggregate AMAE metric. Further, we see in Figure~\ref{fig:AEHierarchical}, that DAER outperforms all baselines on the MAE metric at every coverage greater than  0.197, which corresponds to all cases where fewer than 80.3\% of seeds are rejected. At this crossover point, the MAE is approximately 0.45, meaning in about 1 out of every 222 samples an incorrect answer will be caused by an incorrect seed.
\begin{table}[t]
\centering
\begin{tabular}{|l|l|}
\hline
Method                  & AMAE                         \\ \hline
Random   & $6.17 \pm 1.1e^{-1}$          \\ \hline
Fine Softmax Response   & $3.35 \pm 4.1e^{-2}$          \\ \hline
Fine Entropy            & $3.29 \pm 3.8e^{-2}$          \\ \hline
Coarse Softmax Response & $1.75 \pm 4.6e^{-2}$          \\ \hline
Coarse Entropy          & $1.75 \pm 4.8e^{-2}$          \\ \hline
DAER                    & $\mathbf{1.62 \pm 3.4e^{-3}}$ \\ \hline
\end{tabular}
\caption{AMAE for baselines and DAER on the HSC task (lower is better). Standard error is calculated across five seeding models, and, for DAER, five rejection models.}
\label{tab:HSC_performance}
\end{table}

We also consider the goal of minimizing the number of rejected seeds for a target MAE under the assumption of oracle thresholding. We consider the cases where it is acceptable that 1 out of every 100, 1 out of every 40, and 1 out of every 20 inferences are incorrect due to an incorrect seed, corresponding to acceptable MAEs of 1, 2.5, and 5 respectively. The mean percentage of accepted seeds, as well as the corresponding percent reduction in number of rejected seeds for these cases is shown in Table~\ref{tab:percent_accepted}. Notably, for a target MAE of 5, DAER reduces the number of rejected queries by 23.8\% over the next strongest baseline.
\subsection{Importance of Subgoals}
\label{sec:subgoal_importance}
When defining seed rejection, we proposed two subgoals: \textit{understanding the cause of error} and \textit{understanding the task model response}, which correspond to correctness and regression losses, respectively, in DAER. While we have shown that DAER outperforms the baselines, we have not yet examined the contributions of each subgoal. To do this, we perform three ablations:
\begin{enumerate}[topsep=0pt,itemsep=-1ex,partopsep=1ex,parsep=1ex,itemindent=-2ex]
    \item Correctness: It may be adequate to guess whether or not the seed is a cause of error. To test this, we use the correctness loss alone as the rejection criteria. 
    \item Regression: The way DAER combines its outputs during evaluation is mathematically equivalent to regressing additional error directly. Therefore, we evaluate the value of splitting our loss by training a model to perform regression without the correctness loss. By doing this, we focus solely on understanding the task model's response to the given primary input and seed.
    \item No Seed: While we, in some cases, encourage simplifying the goal of understanding the task model's response by learning which primary inputs are difficult, we would like to ensure that the model does not rely solely on this shortcut. To test if this is the case, we regress the additional error without access to the seed.
\end{enumerate}
\begin{table}[t]
\begin{tabular}{|l|ccc|}
\hline
\multicolumn{1}{|c|}{\multirow{2}{*}{Method}} & \multicolumn{3}{c|}{Target MAE}                                                              \\
\multicolumn{1}{|c|}{}                        & 1                                 & 2.5                                & 5            \\ \hline
Fine Softmax Response                         & \multicolumn{1}{c|}{85.0\%}          & \multicolumn{1}{c|}{67.0\%}          & 24.2\%          \\ \hline
Fine Entropy                                  & \multicolumn{1}{c|}{84.4\%}          & \multicolumn{1}{c|}{64.3\%}          & 22.7\%          \\ \hline
Coarse Softmax Response                       & \multicolumn{1}{c|}{51.8\%}          & \multicolumn{1}{c|}{25.7\%}          & 6.7\%          \\ \hline
Coarse Entropy                                & \multicolumn{1}{c|}{51.7\%}          & \multicolumn{1}{c|}{25.6\%}          & 6.7\%          \\ \hline
DAER                                          & \multicolumn{1}{c|}{\textbf{43.6\%}} & \multicolumn{1}{c|}{\textbf{24.2\%}} & \multicolumn{1}{c|}{\textbf{5.1\%}} \\ \Xhline{4\arrayrulewidth}
Relative Reduction\footnotemark                    & \multicolumn{1}{c|}{15.7\%} & \multicolumn{1}{c|}{5.5\%} & \multicolumn{1}{c|}{23.8\%} \\ \hline
\end{tabular}
\caption{The percentage of seeds which must be rejected for various target MAEs on the hierarchical scene classification task (lower is better), as well as the percent reduction from using DAER over the next-best baseline.}
\label{tab:percent_accepted}
\end{table}
\footnotetext{Calculated: $\frac{\text{Coarse Entropy} - \text{DAER}}{\text{Coarse Entropy}}$} 

We see the results of these ablations in Table~\ref{tab:ablation}, which reveals two interesting phenomena that provide insight into the functionality of DAER: first, in both tasks the correctness loss outperforms the regression loss. Second, even without the seed, understanding the task model's response to the primary input is competitive with some baselines. 

The fact that the correctness loss outperforms the regression loss suggests that classifying seeds as correct and incorrect---understanding the cause of error---is the easier task, and that this rough categorization combined with its implicit confidence is a moderately effective rejection method. However, the fact that it is improved by a conditioned version of regressing the additional error shows us both that eliminating cases where the seed is correct results in an easier regression problem, and that a rejection model trained to solve this regression problem can learn to estimate the task model's response.
\begin{table}[t]
\centering
\begin{tabular}{l|l|l|}
\cline{2-3}
                                                                               & KCVE            & HSC                     \\ \hline
\multicolumn{1}{|l|}{Correctness}                                              & 0.2937          & 1.79 $\pm$ 2.3e$^{-2}$ \\ \hline
\multicolumn{1}{|l|}{Regression}                                               & 1.1633          & 2.05 $\pm$ 1.1e$^{-2}$ \\ \hline
\multicolumn{1}{|l|}{No Seed} &      0.8002           & 2.28 $\pm$ 2.1e$^{-2}$ \\ \hline
\multicolumn{1}{|l|}{DAER}                                                     & \textbf{0.2864} & \textbf{1.62 $\pm$ 3.4e$^{-3}$} \\ \hline
\end{tabular}
\caption{AMAE for DAER and its individual subgoals (lower is better).}
\label{tab:ablation}
\end{table}

\begin{figure}
    \centering
    \includegraphics[width=0.85\linewidth]{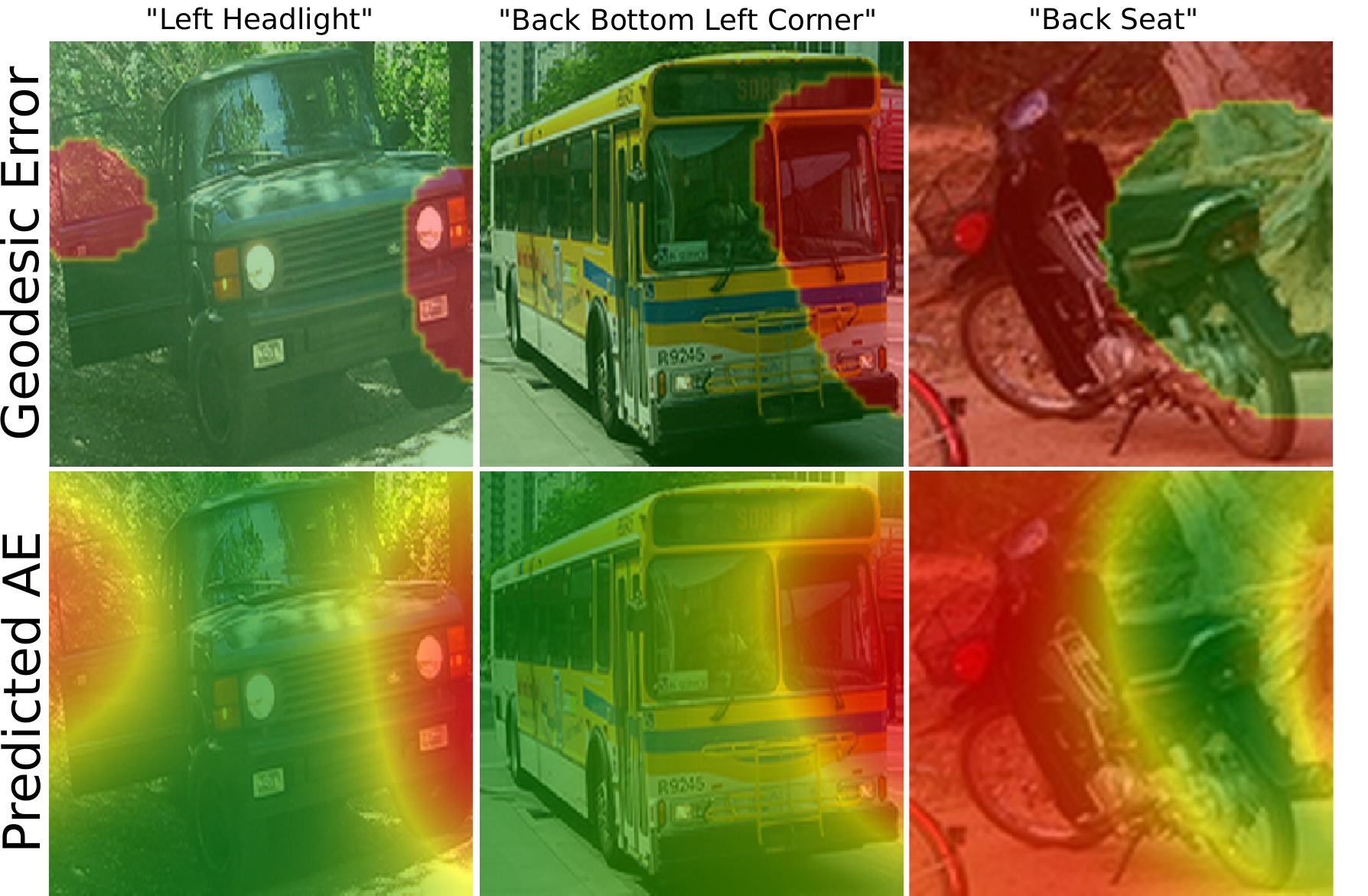}
    \caption{Geodesic error from the task model (top) compared to the additional error prediction from a DAER rejection model (bottom). Error is overlaid from high (red) to low error (green). Predictions are normalized per-image.}
    \label{fig:regressed_AE}
\end{figure}

Further, the fact that performance of a rejection model trained without access to the seed is comparable to baselines on both tasks suggests that it is possible, but not optimal, to perform seed rejection based on the sensitivity of a primary input and task model to the seed. We see why this might be the case in Figure~\ref{fig:regressed_AE}, where the most accurate seed rejection can be performed by regressing the additional error, but rejecting an unknown keypoint on the rightmost image is much more likely to reduce the mean additional error than rejecting an unknown keypoint on the other example images.
\section{Extensions and Limitations}
We have defined the task of seed rejection and associated metrics, AE, MAE, and AMAE.
Using these definitions, we have shown that DAER outperforms baselines due to its novel approach to regressing additional error. In this section, we discuss extensions, limitations, and the implications of a real-world deployment of DAER.

\noindent\textbf{Extensions to Other Common Tasks}
The frameworks of seed rejection and DAER are applicable to a number of tasks which use a seed. In some cases, the extension is straightforward: tasks such as referring expression segmentation~\cite{yu_mattnet_2018,mao_generation_2016, kazemzadeh_referitgame_2014,hui_linguistic_2020} and visual question answering (VQA)~\cite{antol_vqa:_2015,shah_cycle-consistency_2019,wu_improving_2021} are often solved as classification problems, where the primary input is an image and the seed is human-generated text. For these tasks, correct and incorrect seeds can be classified in a manner similar to curated VQA datasets~\cite{prabhakar_question_2018,gurari_vizwiz_2018} while additional error can be regressed directly. The additional error derived metrics can likewise be used in a straightforward way. 

In tasks where the seeds are non-homogeneous---such as using differing hierarchy levels for hierarchical scene classification or choosing between keypoint or dimension line~\cite{lemmer_crowdsourcing_2021} annotations for human-in-the-loop viewpoint estimation---the formulation holds, but the definition of seed as containing multiple, inconsistently used, input modes leads to a challenging set of architectural challenges.

For tasks such as single-target visual object tracking~\cite{bertinetto_fully-convolutional_2016,kristan_novel_2016,fan_siamese_2019} and video object segmentation~\cite{perazzi_benchmark_2016,khoreva_learning_2017,caelles_one-shot_2017}, there is a meaningful challenge in defining the gold-standard seed: while relevant datasets provide a gold-standard seed, there is no guarantee that it will be the best performer nor that the seeding method will tend to provide this gold-standard seed.  Further exploration of these interesting questions is beyond the scope of this paper.

\noindent\textbf{Implications of Deployment}
The goal of seed rejection is to reduce the potential impact of incorrect seeds at inference time. In doing so, it reduces the number of seeds required for a target accuracy, thereby lowering the cost of deployment and making such artificial intelligence solutions more broadly accessible. Since the ultimate goal of seed rejection is to correct inferences that are already incorrect for a fixed task model, it is unable to increase the impact of any bias or failure mode over a deployment that does not utilize a rejection model although, like all models, a DAER rejection model is subject to its own failure modes.

One notable exception to this is if DAER is extended to dataset curation, either through active learning~\cite{yoo_learning_2019} or by removing training data that degrades model performance~\cite{shah_choosing_2020}. Since such an application would establish a bidirectional dependency between the task and rejection models (\ie, the task model trains the rejection model which trains the task model), the ultimate point of convergence is unclear, and may amplify biases or blind spots. As such, we do not recommend a direct extension of the findings of DAER to dataset curation without a thorough investigation of this phenomenon.

\section{Conclusion}
In this work, we introduced the novel problem of seed rejection, addressing for the first time the impact of individual incorrect seeds on a model's performance. In problem-agnostic terms, we introduce the evaluation metrics of additional error (AE), mean additional error (MAE), and area under the mean additional error curve (AMAE), and designate two meaningful subgoals: understanding the cause of error, and understanding the task model response. These subgoals motivate the  Dual-loss Additional Error Regression (DAER) method, which we show can reduce the number of required re-annotations for a target MAE by over 23\% compared to the best-performing baseline.

\small{\paragraph{Acknowledgements} Toyota Research Institute ("TRI") provided funds to assist the authors with their research but this article solely reflects the opinions and conclusions of its authors and not TRI or any other Toyota entity.}

{\small
\bibliographystyle{ieee_fullname}
\bibliography{egbib}
}

\end{document}


\title{Supplementary Material\\ \large
Ground-truth or DAER: Selective Re-query of Secondary Information}


\maketitle
\ificcvfinal\thispagestyle{empty}\fi

\section{Architecture and Training Details}
\begin{figure}[t]
    \centering
    \includegraphics[width=\linewidth]{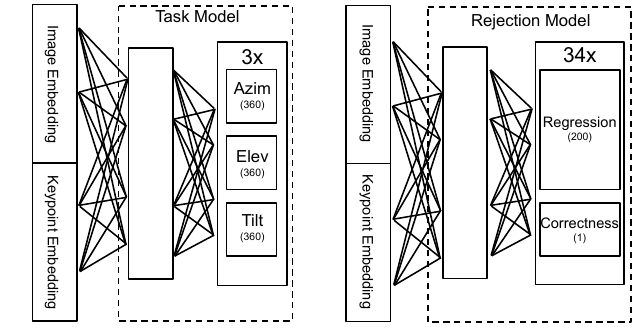}
    \caption{Prediction layers for the KCVE task (left) and rejection (right) models, which accept the keypoint and image embeddings from~\cite{szeto_click_2017}.}
    \label{fig:prediction layers}
\end{figure}
\subsection{Keypoint-Conditioned Viewpoint Estimation}
\paragraph{Architecture} The Click-Here CNN (CH-CNN) architecture consists of separate, mostly independent, branches to process the image and keypoint. The features produced by these branches are concatenated and passed through two linear layers to produce the desired output. As this architecture has been proven capable of integrating keypoint and image data, we use it not only for the task model, but also as the backbone of the rejection model with the output layers shown in Figure~\ref{fig:prediction layers}. Further information on the base architecture is available in the original work~\cite{szeto_click_2017}.

The output of the task model (Figure~\ref{fig:prediction layers}-left) is of size 3x3x360, consisting of three vehicle classes (car, bus, motorbike), three angles (azimuth, elevation, tilt) and 360 potential angle values. The output of the rejection model (Figure~\ref{fig:prediction layers}-right) is of size 34x(200+1), consisting of 34 potential keypoint classes, 200 binned outputs per keypoint class to regress the additional error, and one output per keypoint class to estimate the correctness.

\paragraph{Training}
The rejection model is trained in two phases. In the first phase, it is trained on a combination of rendered~\cite{szeto_click_2017} and real~\cite{xiang_beyond_2014} data. Candidate seeds are generated by randomly selecting an x-y location on the image. An Adam optimizer~\cite{kingma_adam_2017} is used with learning rate 1e$^{-4}$ and early stopping is performed on the validation loss with a patience of 5 epochs.

In the second phase, the rejection model is trained exclusively on the PASCAL3D+ dataset~\cite{xiang_beyond_2014}. The same optimizer settings are used, however the one-hot additional error target is softened by convolving with a Gaussian kernel with standard deviation 3. Early stopping is performed on validation loss with a patience of 100 epochs.

Regression and correctness ablations use the same training procedure, where back propagation is only performed on the appropriate loss.  For the no seed ablation, a tensor of zeros is given to the rejection model in place of the keypoint map, and no further modifications are made to architecture or training.
\begin{figure}[t]
    \centering
    \includegraphics[width=\linewidth]{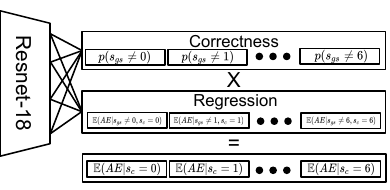}
    \caption{Our rejection model architecture and output format for the HSC task. Each potential candidate seed is given two outputs, which are multiplied to estimate the expected additional error.}
    \label{fig:hsc_arch}
\end{figure}
\begin{figure*}[t]
    \centering
    \includegraphics[width=.95\linewidth]{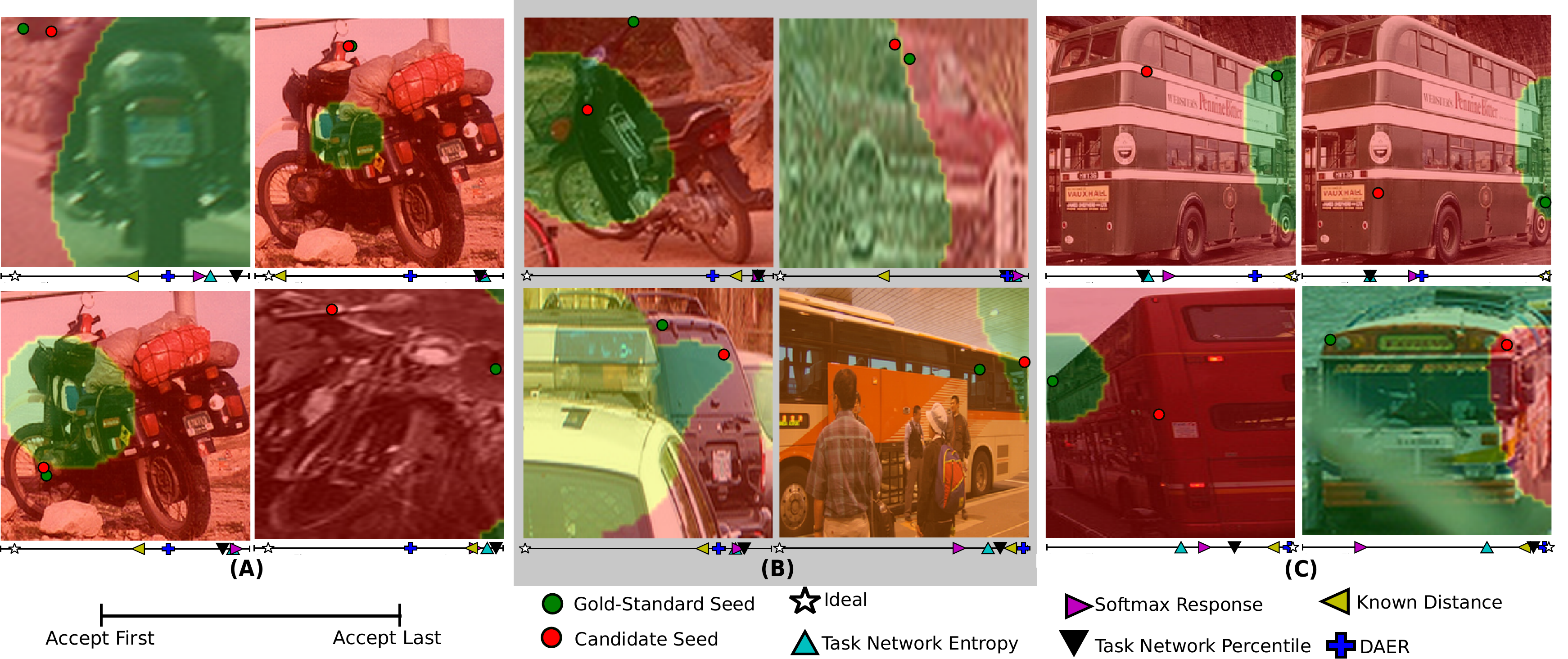}
    \caption{Quantitatively chosen KCVE heatmaps.  The gold-standard seed is shown in green, the candidate seed is shown in red, and a red-yellow-green heatmap gives the additional error for that keypoint click. Methods closer to the white ``ideal'' star are better for that example. (A) The four cases where the gold-standard seed provided the worst absolute performance. (B) The four cases where the candidate seed improved upon the gold-standard seed the most. (C) The four cases with the highest additional error.}
    \label{fig:quant_examples}
\end{figure*}
\subsection{Hierarchical Scene Classification}
The architecture and output layers used for the hierarchical scene classification task are shown in Figure~\ref{fig:hsc_arch}. As a backbone, we use a ResNet-18 which has been pretrained on ImageNet~\cite{deng_imagenet_2009}, and truncate the output to 2 elements per seed class (14 total). Seven of these outputs---the correctness outputs---are trained using a cross-entropy loss to determine 
\begin{align}
p(s_{gs} \neq \text{class} | x) = 1 - p(s_{gs}=\text{class}|x)\enspace,
\end{align}
\noindent while the other seven are trained using a binary cross entropy to find 
\begin{align}
\mathbb{E}(\myAE | x, s_{gs}\neq \text{class}, s_c=\text{class})\enspace .
\end{align}

\noindent The model is trained for 50 epochs with learning rate 1e$^{-5}$ and the model with the best validation AMAE is used for evaluation. The correctness-only rejection models, regression-only rejection models, and seeding models, are trained identically using only the appropriate outputs, except the learning rate is increased to 1$e^{-4}$ and accuracy is used in place of AMAE to select the best seeding model.

Since the individual outputs in this architecture correspond to different seeds, the blind ablation was performed by reducing the output to a single value that regresses the additional error regardless of the input seed.


\section{Quantitatively Chosen KCVE Examples}
In the main text, we use qualitatively chosen examples to illustrate characteristics of the task and rejection models. Here, we show additional examples that were selected using quantitative criteria on our crowdsourced keypoints: the four cases where the gold-standard seed resulted in the highest geodesic error (Figure~\ref{fig:quant_examples}-A), the four cases where the candidate seed improved upon the gold-standard seed the most (Figure \ref{fig:quant_examples}-B), and the four cases with the highest additional error (Figure~\ref{fig:quant_examples}-C).

Figure~\ref{fig:quant_examples}-A shows four cases where  the performance of the candidate seed is poor but should be accepted, as we would expect a worker to continue returning seeds that are near the ``target'' gold-standard seed even though it won't result in better performance. We see that DAER outperforms baselines that do not have prior knowledge of the gold-standard seed location in all four cases by accepting these instances earlier.

In Figure~\ref{fig:quant_examples}-B, where the candidate seed improves upon the gold-standard, the rejection model must understand that despite returning a different answer than the gold-standard, the candidate seed does not make performance worse. Given the depth to which the rejection model must be able to understand the task model to make this distinction, it is unsurprising that no method clearly outperforms the others on these four samples.

Figure~\ref{fig:quant_examples}-C answers the most intuitive question of seed rejection: how well does a rejection model reject seeds with high additional error? We see that while using an oracle measure of distance is best in some instances, DAER outperforms all baselines that do not have prior knowledge of the correct answer in all four cases by accepting these instances last.





\begin{figure*}[t]
    \centering
    \includegraphics[width=.72\linewidth]{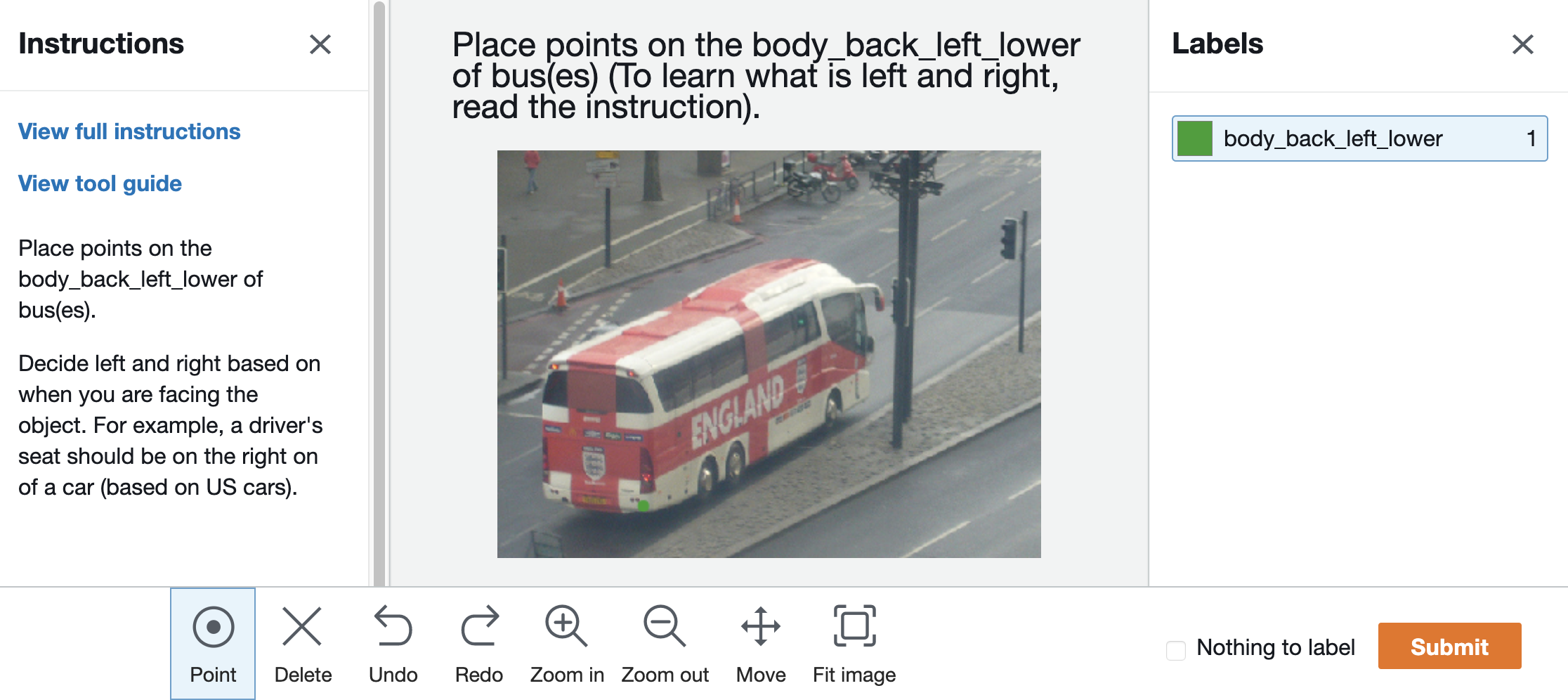}
    \caption{The interface provided to crowd workers for crowdsourcing keypoint clicks.}
    \label{fig:annot_interface}
\end{figure*}



\section{Crowdsourcing Keypoint Clicks}

Keypoint annotations are collected from US-based annotators using the interface shown in Figure~\ref{fig:annot_interface}. The worker is shown an image containing one or more vehicles, and is asked to click all instances of a specific keypoint class. If an annotator responded that the keypoint class wasn't present, we provided the query to another annotator up to two additional times. If all three annotators responded that the keypoint class wasn't present, we assumed the gold standard was incorrect or too difficult, and removed it from the evaluation.

To match the annotated keypoints with the corresponding verified gold-standard keypoint from PASCAL3D+, we use a three-step process: first, we associate all keypoints to vehicle crops which contain them. Next, we match these keypoints to the gold-standard keypoint of the same class in that vehicle crop. Last, if a vehicle crop contains multiple candidate keypoints of the same class, we select the one that is nearest to the gold-standard keypoint. Using this process, we receive annotations matching 6,042 of the 6,593 gold-standard keypoints.

Analyzing the distribution of matched keypoints, we found that 40\% of keypoints were within 5 pixels of the matching gold-standard and 57\% were within 10 pixels of the matched gold-standard keypoint. 
We further found that 6.3\% (381) of keypoints cause additional error, while 1.3\% (81) cause more than 5$^\circ$ additional error, and 0.5\% (30) cause more than 150$^\circ$ additional error.
\begin{table}[h]
\centering
\begin{tabular}{|l|l|}
\hline
Percentile & AMAE   \\ \hline
70\th      & 0.3472 \\ \hline
80\th      & 0.3092 \\ \hline
90\th      & 0.3303 \\ \hline
\end{tabular}%

\caption{AMAE at various sampling percentiles.}
\label{tab:percentile-sweep}
\end{table}
\section{Evaluation of Sampling Method by Percentile}
For the KCVE task, we consider a sampling-based baseline in which 10,000 samples are taken from the CH-CNN output distribution, and the sample at the $n$\th\hspace{.25mm} percentile distance from the mean is used as the scoring function. We present the results for the 70\th, 80\th, and 90\th\hspace{0.25mm} percentile in Table~\ref{tab:percentile-sweep}, justifying our choice of the 80\th\hspace{.25mm} percentile as our baseline.

\section{Per-Fold KCVE Results}

Table~\ref{tab:per-fold} shows the per-fold AMAE of the various rejection models on the KCVE task. We see that no single method performs best across all folds, but DAER is the most consistent: DAER does not perform worse than 25.3\% above its mean on any fold, while the corresponding number for the best baseline (sampling percentile) is 80.4\%. 

As each baseline only addresses one of the described subgoals (e.g., distance only finds the cause of error and sampler only understands model response), this suggests that some folds contain more instances of one source of error, and again highlights the importance of the subgoals described in the main paper. While focusing solely one subgoal allows baselines to perform well on folds where that source of error is more frequent, DAER's understanding of both subgoals leads to more consistent and overall better performance.

\begin{table}[h]
\resizebox{\linewidth}{!}{%
\begin{tabular}{l|l|l|l|l|l|l|}
\cline{2-7}
                                       & \multicolumn{6}{c|}{Fold}                      \\ \hline
\multicolumn{1}{|l|}{Method}           & 1     & 2     & 3     & 4     & 5     & Mean   \\ \hline
\multicolumn{1}{|l|}{Softmax Response} & 0.561 & 0.999 & 0.167 & 1.430 & 1.496 & 0.9306 \\ \hline
\multicolumn{1}{|l|}{Distance}         & 0.419 & \textbf{0.254} & 0.147 & 0.757 & 0.405 & 0.3964 \\ \hline
\multicolumn{1}{|l|}{Entropy}          & 0.325 & 0.556 & 0.112 & 0.421 & 0.353 & 0.3534 \\ \hline
\multicolumn{1}{|l|}{Sampler}          & \textbf{0.292} & 0.558 & 0.125 & 0.370 & \textbf{0.201} & 0.3092 \\ 
\Xhline{4\arrayrulewidth}
\multicolumn{1}{|l|}{Correctness}             & 0.312 & 0.343 & 0.118 & 0.414 & 0.282 & 0.2937 \\ \hline
\multicolumn{1}{|l|}{Regression}             & 2.342 & 0.274 & 1.102 & 0.992 & 1.107 & 1.1633 \\ \hline
\multicolumn{1}{|l|}{No Seed}             & 0.852 & 0.778 & 0.480 & 1.003 & 0.889 & 0.8002 \\ \hline
\multicolumn{1}{|l|}{DAER}             & 0.322 & 0.307 & \textbf{0.109} & \textbf{0.335} & 0.359 & \textbf{0.2864} \\ \hline
\end{tabular}%
}
\caption{Per-Fold AMAE on the KCVE task. Baselines are above the thick line, ablations and DAER are below. The best performer per-fold is shown in bold (lower is better).}
\label{tab:per-fold}
\end{table}
{\small
\bibliographystyle{ieee_fullname}
\bibliography{egbib_2}
}